# I-POST: Intelligent Point of Sale and Transaction System

Farid Khan
Department of Computer Science
Kennesaw State University
Mariatta, GA
fkhan14@students.kennesaw.edu

*Abstract*— We propose a novel solution for the cashier problem. Current cashier system/Point of Sale (POS) terminals can be inefficient, cumbersome and time- consuming for the users. There is a need for a solution dependent on modern technology and ubiquitous computing resources. We present I- POST (Intelligent Point of Sale and Transaction) as a software system that uses smart devices, mobile phone and state of the art machine learning algorithms to process the user transactions in automated and real time manner. I- POST is an automated checkout system that allows the user to walk in a store, collect his items and exit the store. There is no need to stand and wait in a queue. The system uses object detection and facial recognition algorithm to process the authentication of the client and the state of the object. At point of exit, the classifier sends the data to the backend server which execute the payments. The system uses Convolution Neural Network (CNN) for the image recognition and processing. CNN is a supervised learning model that has found major application in pattern recognition problem. The current implementation uses two classifiers that work intrinsically to authenticate the user and track the items. The model accuracy for object recognition is 97%, the loss is 9.3%. The model accuracy for facial recognition is X% and the loss is Y%. We expect that such systems can bring efficiency to the market and has the potential for broad and diverse applications.

*Keywords—object detection, facial recognition, Convolution neural network, Tensorflow*

## I. INTRODUCTION

The United States offers the largest consumer market with a GDP of $20 trillion dollar and population of 325 million [1]. It is estimated that a regular customer on average spends 41 minutes in a shopping trip (including waiting in cashier line). This means over 53 hours per year are spend in the grocery store [2]. This place a high overhead on the providers and high time consumption for the customers. We expect that the technology age can provide solutions. There are two revolutionary technologies that will change the consumer/ retail and logistic industry. These are the following:

1. Smart devices: Smart Devices/ IoT are devices that uses hardware and software controllers to manipulate the environment and communicate with other devices or the Internet- based services [3]. Smart device use sensors to retrieve input. Then use limited computing power to perform calculation and impact the environment through actuators. Further the device can send/ retrieve data from servers or other utilities. For instance: the smart door lock can detect the presence of the owner phone and send signal to open the door. If the temperature rises above a threshold, then smart thermostat send ping to window sensors to open the window. These devices have the potential to provide home automation and digital convenience.

2. Machine Learning: Tom Mitchell (pioneers in machine learning) stated that machine learning is a computer program that is designed to solve task T, while the program learns from Experience E. The task is measured against a performance metric P [4]. For example: the computer program task is to perform binary classification. The program enacts and gain an output. The output is compared to Performance metric (is the output equal to actual value or not), which allows the program to learn if it calculated the correct or incorrect answer. This can be stated as to be supervised learning (the program learns on data with labels).

Machine Learning is a sub- field in the artificial intelligence that deals broadly with the application of intelligent systems. For instance: auto recommender systems, machine vision, data mining, Natural Language Processing, Automated Driving etc. These two technologies (in conjunction with network security) has played a major role in transforming the retail market. There are number of applications in smart cities/ smart stores expected in hindsight. The four fundamental technologies that allowed a paradigm shift in the retail market are indoor navigation system, augmented reality headsets, facial recognition algorithms and interactive digital signage [3].

### A. Research Problem

The problem is to find novel ideas and solutions for cashier less payment technology. Cashier- less pay means that the user can pay on the go and there is no need for human employee at the point of sale. As aforementioned, the rise of Artificial intelligence and smart devices will impact the user purchase mechanism. We expect that this research can lead to practical application of machine learning algorithms and will spur further research in the projected domain.

*B. Purpose of the Study*

The purpose of this work is to design an intelligent system that can perform cashier- less payments. The system uses machine vision to detect the objects and to authenticate the user with facial recognition. Facial Recognition is the ability to recognize a person through image data input [4]. The back-end server handles the processing and update for the client side. We expect that such software solutions to gain broad support as smart stores become mainstream.

*C. Audience*

This research is open to all readers who are interested in understanding about machine learning, pattern recognition and smart services. The target audience and engineers who would like to learn to develop intelligent systems for applications and business executives who covet to stay up to date with cutting – edge technology and to understand the market potential of I-POST and related software systems.

*D. Contribution*

Our contribution is to research the potential of machine vision and intelligent theory and applicability. We designed a mechanism to track the object source/ location and to authenticate the user with facial features. At time of exit, the user items are total, and the receipt is presented online. We provide experimental results and implementation methods to illustrate our I- POST software system.

*E. Motivation*

Due to large computing resources, spurs of smart sensors and breakthrough in the field of artificial intelligence, has allowed us to apply technology to solve many society-oriented problems. This paper concerns with retail stores and automation of cashier systems. These reasons are the motivation to the work and the information is provided in a comprehensive manner.

*F. Paper Organization*

The paper organization includes a brief history of transaction services, theory of machine learning, convolution neural network mechanism, implementation method and experimental results. We further expound on the limitation of the system which can open future research in the specified domain.

## II. RELATED WORKS

*A. Background*

A retail store is a place where goods, services or commodities are sold to the public. The four categories of retail are: long- term goods (car, furniture), short -term (clothes), Food / consumption items and art (music albums, art painting) [5]. The retail industry is considered the largest employer of US in the public sector, which accounts to 10% of the employment [6]. A retail store (in physical) uses a Point of Sale (POS) terminal. POS is the place where items are sold to the client, client makes the payment and a receipt is printed for the client. As aforementioned, this process can be long, tiresome and overhead for the client and the host as well. This can be phrased as the cashier/ transaction optimization problem.

With the advent of cheap electronic, high computing power and sophisticated algorithms, the transaction problem potentially has a solution. It is a cashier- less technology. A case study of such a technology is found in Amazon smart stores. Amazon has been at the fore front of this technology and opened their smart store in 2018. The service is called Amazon Go. The system is designed as such to use user mobile phone for authentication and registration purposes. The user first downloads the Amazon Go app. Whenever the user enters the store, he slides his phone through the door sensors. This triggers the authentication procedure. Once authorized, the user can continue to shop in the smart store. The store is equipped with multiple cameras mounted on the ceiling that covers the store. The images retrieved are send to personal server to perform calculation about the object location and information [7]. *Our current research is limited on their implementation methods and camera coordination, because Amazon hasn't published their work or made it public.* Further, Amazon Go doesn't use facial recognition (the ability to recognize an induvial through image data input) for payment. Rather they use an intricate algorithm to measure the whereabouts of the user (for instance: hand motion, body coordinates).

The advantage of the Amazon Go system is the convenience to the user, efficiency (the user doesn't have to wait in line), cost effective (as less resources are required) and high productivity. The disadvantage is still in the accuracy and the speed of the system. The technology is still in infancy and will require more test and evaluations to see the full – scale impact of cashier -less solutions [7].

*B. CNN*

Artificial Intelligence has seen momentous advancement in the past decade. The original works dates to 1943, when Walter Pitts and Warren McCulloch proposed the idea of neural networks. They showed the first mathematical model that described the functioning of neural networks.

As aforementioned, Machine learning is a sub- field within the AI domain (other sub- fields include cognition, neuroscience, linguistics, anthropology etc.) which pertains to pragmatic applications of the computer. It is to make a computer program learn a task and optimize it against a set benchmark.

All machine learning can be categorized, generally, in two learning types: supervised and unsupervised. Supervise learning is in which the input is provided with the appropriate output label. In unsupervised, the input is not provided the any output parameters.

Convolution Neural Network (CNN) are supervised learning and deep neural networks (contain many layers of hidden layers between the input and output) [8]. CNN are derived through understanding the workings of human cerebral cortex (which plays a major role in human ability to see and navigate). CNN architecture has vital characteristics that makes the system applicable for image/ vision recognition problems.

The earliest research in the field of information processing through visual stimuli was carried by David H. Hubel and Torsten Wiesel (1960s) [8]. There finding was that human brain uses two cells that play a role in our ability to see. Simple cells are in front layer of visual processing and respond to the stimuli in receptive field. Complex cell takes the input from the simple cell to calculate the output [9]. This was a building block to ConvNet architecture. In 1980, Kunihiko Fukushima, Japanese researcher, introduced the concept of convolution layers and downsampling layers (the description of the terminology is presented in CNN Glossary) [8]. In 1989, Yann LeCun, Senior Researcher at Facebook, was able to use CNN architecture to predict hand – written numbers [9]. The paper applied backpropagation (a technique to recalibrate the weights according to the loss function) on the hidden layers. The system outperformed the previous models by a margin. The ConVnet model was later applied by bank companies to read checks and payments. To provide higher accuracy, the need of deep neural network (neural networks with multiple layers of interconnected neurons) arise. In 2012, Alex Krizhevsky, used AlexNet (A CNN contain 8 layers), to win the ImageNet Large Scale Visual Recognition Challenge. The model achieved only 15% error rate [10]. In 2015, the performance measure bar was elevated further by Microsoft 100 layer CNN on the same challenge [11]. This shows that will deeper/ complex networks and GPU processing power, unfathomable things can be accomplished through an electronic brain.

These are the following definition and information on the terminologies used in CNN Architecture:

[1] Convolution Layer: It is the foremost layer in the CNN architecture. It collects important features from input image. This requires a filter (a matrix that performs dot product on the input) which gives us the feature map (matrix that shows the extracted information from input). Convolution is a mathematical concept which states that when two functions are added, we acquire a new set of function [12].

[2] Max- Pooling Layer: Pooling is a method to extract more valuable information form the convolution operation. This is achieved through reducing the dimensionality of the data and retain the spatial invariance (the target image may slightly change without losing it features). Max – Pool is to further extract the important features form the prior layer. For example:

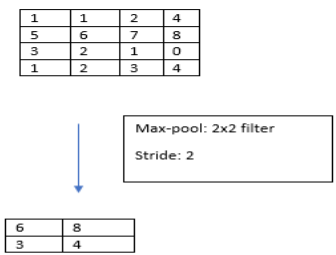

Figure02: pooling operation example

[3] Shift invariance: when the input is shifted by a degree, then proportionally the output is shifted [13]. This shows, in simple terms, if the image is rotate or deformed, the prediction won't change. As the features (domain knowledge) remains intact. In mathematical notion:

$$f(x(t) = g(y(t))$$
$$f(x(t+r)) = g(y(t+r))$$

We can see that if the input is shifted with respect to time, then the output finds the exact shift as well. This keeps the features to scale.

[4] Feed- Forward Neural Network: This is one- way function, in which the data travels only from input to output. It doesn't have similar model as recurrent neural net, which contains cycles and allows the output to be used as a feedback mechanism [14].

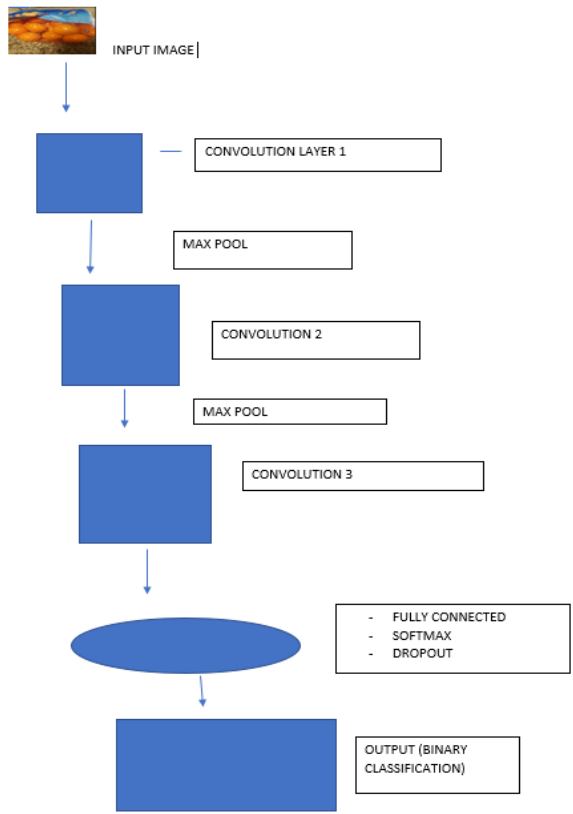

Figure 01: CNN – Architecture Description

[5] Fully – connected Neural Net: This is part 3 of the CNN architecture after the data has been processed through the hidden layers.

[6] SoftMax function: this is an activation function. An activation function is derived from neuron mechanism. Neurons either fire or no. binary operation is performed. Similarly, an activation function acts as a threshold. Softmax is a function that converts the input values into a probability distribution which is used on the feature space [15]. The higher the probability of a feature, corresponds to the degree of certainty. The following is the equation:

$$y(x) = \frac{e^{x(i)}}{\sum e^{x(i)}}$$

[7] Dropout: It is a mechanism to stop the model form overfitting the data [16]. Dropout randomly selects nodes which are removed from the learning during training set. This is called dropping units. During epoch runs, this forms into thin network which are then sum in the outcome. This halts overfitting form occurring. All the runs are then added to achieve the designated output.

### III. IMPLEMENTATION

#### A. Techniques Used in the study

the current work is to propose a solution for the cashier problem. We created a system called I - POST, which uses the facial recognition and object detection mechanism to charge the user. The input parameters will be phone information, facial features and objects in basket. This information is then processed in the back end to charge the customer.

The following are the requirements for the I – POST deployment strategy:

The implementation model requires two classifiers, regular RGB- camera, raspberry Pi, server and a configured back-end (Payment handling through Web API calls).

There are three phases for implementation trajectory:

1. classifier to be able to work on input data.

[results achieved and presented below]

Phase 1 consist of the two classifiers. there is an object detection classifier and face detection classifier. Both models work on CNN architecture. We use Keras library for implementation. This implementation details are explained in the paper.

2. configure the dataset for transaction purposes.

[We are currently conducting test on this configuration]

3. demo- based model

[work in progress]

The project is the one in which the remaining in the scenario. To be judges and this is the one

#### B. Proposed Method

The model we used has two parts. the description are presented on the following:

1. Classify the object items and face detection through the classifiers.

The object detection used the python and Keras package. We used tensorboard as a package to achieve the graphs and the visualization.

2. Face recognition and detection:

We applied Python package (face recognition) to perform the task of facial classification on the train dataset. We used a public figure and broadly available data, so we chose US President Donald Trump, for the test and validation purpose.

3. Raspberry Pi (Demo):

Raspberry Pi can provide two benefits to the system. It can allow the classifier to work together (food and face) and it can be performed in real – time. We expect to provide further details as currently we perform test.

We designed a high-level theoretical model that is able to capture the data flow, information retrieval and the back-end configuration:

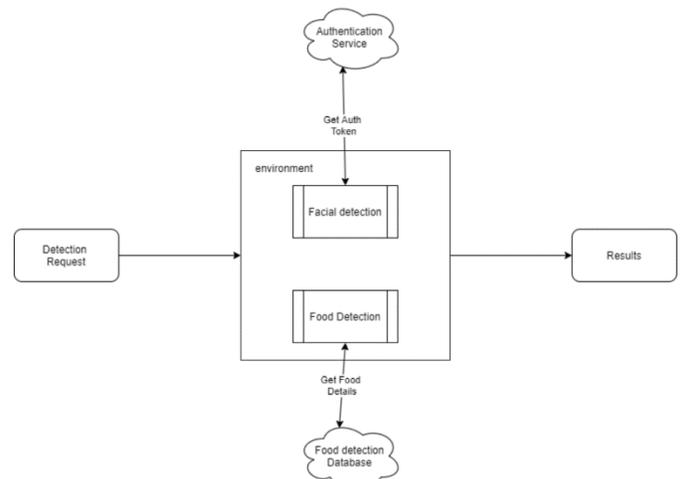

Figure02: This is the high- level view of the I-POST. The detection request is the input or trigger for authentication protocol. The user receives an authentication token which allows the user to shop. The face and food detectors are initiated. At checkpoint, the final result is send to the back-end to perform transaction protocol. The Auth token expires at exit.

*C. Dataset*

The dataset that we used for food object detection is provided through Kaggle competition [5]. The dataset is 5GB and contains 10000 images of food. The following images are used for train and test purposes:

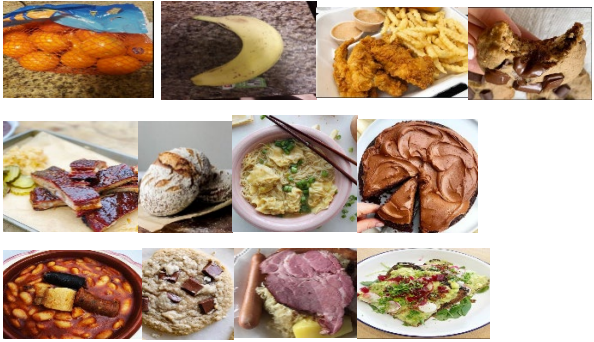

Figure 03: Images of training set for the object detection. These images are snapped by the authors. The images from top-left: pack of oranges, banana, hamburger/ fries, chocolate cookie, BBQ ribs, sourdough bread, Vietnamese soup, Chocolate Cake, Potato soup, Chocolate Cookie, Ham/ Hotdogs, Veggie Platter.

In order to perform and test the facial recognition system, we decided to use a public entity (for test) as there is ample dataset and quick results can be achieved. The following are the images we used for the train/ test purpose.

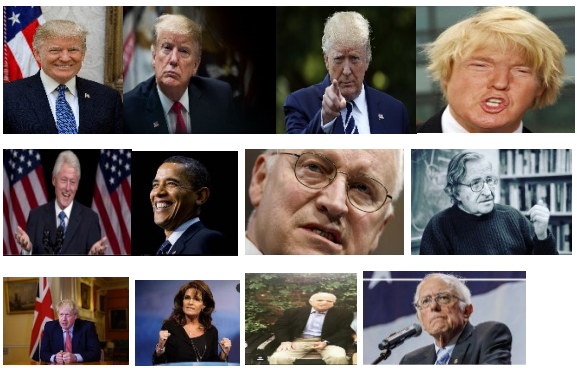

Figure04: This is the dataset we manually created for facial detection scenario. We focused on public personalities and politicians for training set purpose. The test set uses Donald Trump images, as shown in the results section below. The description of the pictures are the following: Donald Trump (Top Left – google

To further the face recognition test, we collected pictures of other famous personalities. These are the following:

## IV. EXPERIMENT AND RESULTS

| Parameters | Name/ Type | Result |
| --- | --- | --- |
| Optimization | adam | N/A |
| Loss Function | Binary- cross entorpy | 0.619 |
| Metrics | Accuracy | 0.78 |

Table1: This is the summary of the implementation of object detection that was performed on a Kaggle food dataset (100000 images), which is present in the links.

| Parameters | Name/ Type | Result |
| --- | --- | --- |
| Optimization | adam | N/A |
| Loss Function | Triplet Loss Function | 0.80 |
| Metrics | Accuracy | 0.96 |

Table2: this is the summary of the implementation of the face detection mechanism. We collected multiple images of famous personlities (through web search/ web crawling). The measurements are presented in the table.

These are the following result we reached after performing test on the food item list. Orange line represent the train set while blue represents the validation set. This experimental run is performed as binary classification of two item (omelette and donut).

The face detection is performed on keras implemetnation of CNN. Further we also tested with SVM classifier. We used dataset of images 250 x 250 dimensions. The triple loss function is applied. The learning is taking place as the simialrity index gets higher. This means that as more similar the feature scales or the distance between the attributes are lower, the likelihood is the classifier knows the image. The mechanism doesn't include dynamic or real – time classification. We were able to achieve good result on the static input test.

The following are the graphs:

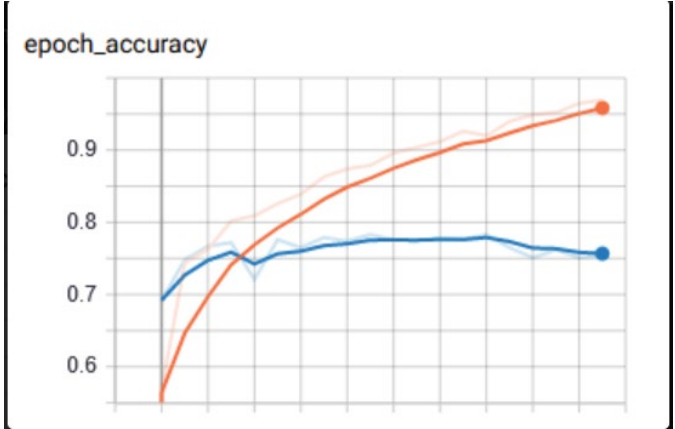

Figure05: Accuracy rate increases with number of epochs. The more the number of epochs and fint- tuning of parameters can achieve high learnig and accuracy results. We used CNN and SVM model for face detection/ recognition.

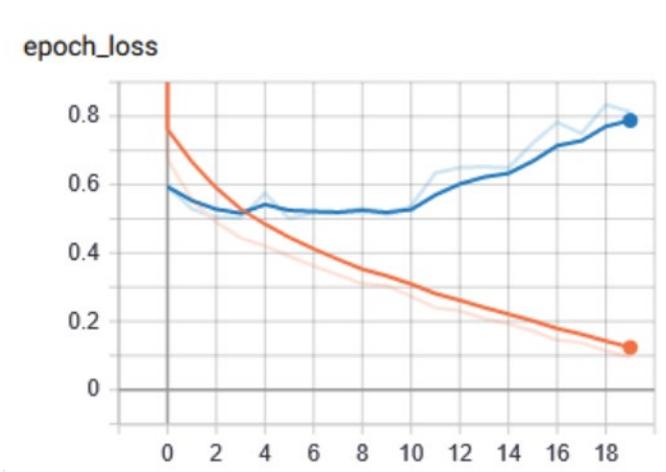

Figure06: Decrease in error rate with respect to the number of epochs

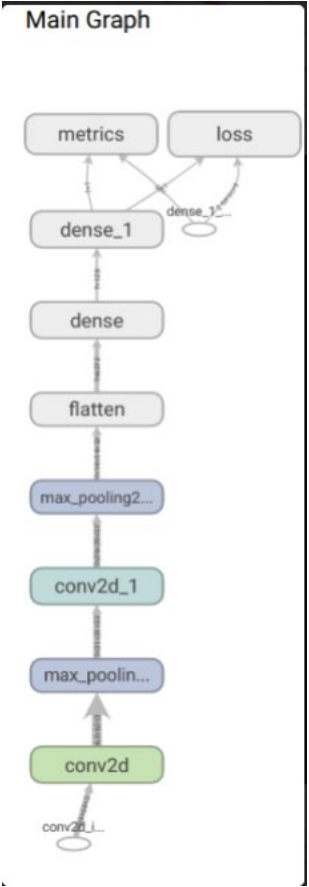

Figure05: Convolution neural Network Design Architecture. It contians 3 convolution layer and 3 max pooling layers. This information is then processed in Full Neural Network. The optimiztion and loss funciton are applied. The output is retireived in binary values.

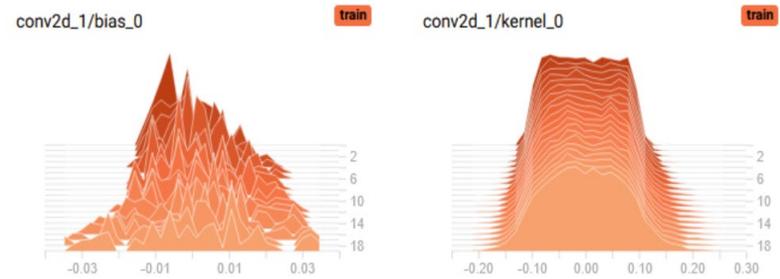

Figure06: Relation of kernel with the training set. The learning rate is dependent on the hyperparameter settings and the number of datasets. The peak in the graph represents high correlation.

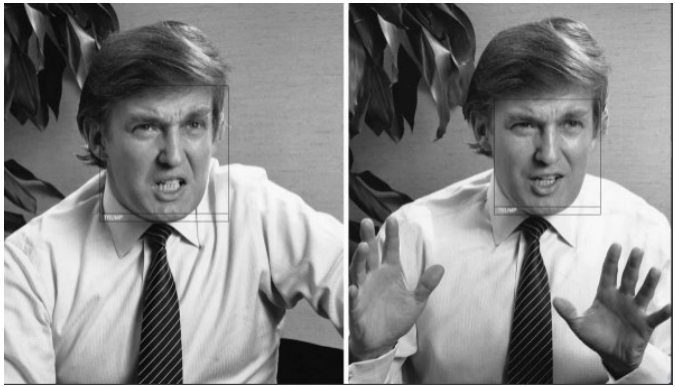

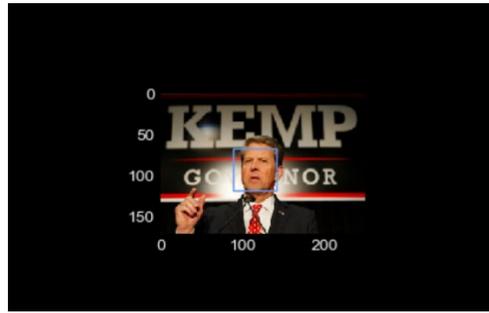

Figure10: The localization of the test set is depicted. The feature is then extracted and compared with learn parameters.

Figure07: This illustrates the facial classifier to predict the image as of Donald Trump. The dataset uses 250 x 250 pixel , rgb, image files (jpg or jpeg format are used). High accuracy is reached depending on the quality and the localization of the test image set.

We are currently working on a mechanism to capture images in real- time, which can substantially increase the use and reliability of face detection system. We also need to fix the issue of making the two classifiers work together through Keras library. We expect to apply the I – POST model through Arduino microcontroller set, which can enhance the robustness and the data handling of the system. This research has wide application and potential to impact the retail, tourist and marketing industry. In future, the system can be provided through a Web API call. This can allow wide deployment, copyright protection and central control on the software system. Further test and evaluation is required to increase the accuracy and time sensitive performance.

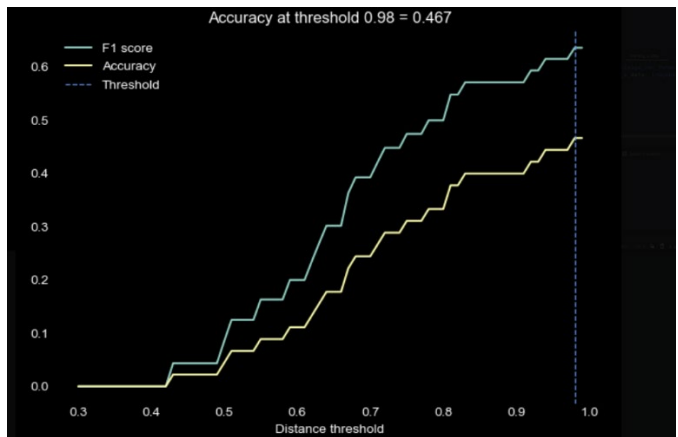

Figure08: the graph depicts the learning of the face classification. F1- score is the test accuracy measurement. The correlation of feature is the domain of accuracy metric.

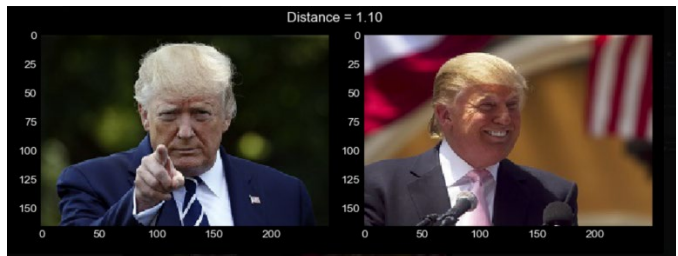

Figure09: the feature similarity index measurement. As the classifier works, it extracts the important feature. The shorter the distance between the error loss, higher the similarity of the test features are displayed.